\documentclass{article}
\usepackage{spconf,amsmath,graphicx}
\usepackage{color}
\usepackage{times}
\usepackage{epsfig}
\usepackage{stfloats}
\usepackage{algorithm}
\usepackage{algorithmic}
\usepackage{multirow}
\usepackage{booktabs}
\usepackage{bbding}
\usepackage{float}
\usepackage{subfigure}
\usepackage{paralist}
\usepackage[misc]{ifsym}
\usepackage{amsfonts}
\usepackage{textcomp}
\usepackage{enumitem}
\usepackage{threeparttable}
\usepackage[numbers]{natbib}
\usepackage{array}
\usepackage{url}
\usepackage{arydshln}
\usepackage{caption} 
\usepackage{makecell}
\usepackage{graphicx}
\usepackage{subfigure}

\title{MPASNET: Motion Prior-Aware Siamese Network for Unsupervised Deep Crowd Segmentation in Video Scenes}

\name{Jinhai Yang, Hua Yang\sthanks{Corresponding author: hyang@sjtu.edu.cn. 
This work was funded by National Natural Science Foundation of China (NSFC, Grant No. 61771303), Science and Technology Commission of Shanghai Municipality (STCSM, Grant Nos. 19DZ1209303, 20DZ1200203, 2021SHZDZX0102), and SJTU Yitu/Thinkforce Joint Laboratory for Visual Computing and Application.
}
}
\address{Institute of Image Communication and Network Engineering, 
Shanghai Jiao Tong University, China
\\Shanghai Key Lab of Digital Media Processing and Transmission, Shanghai, China
\\
MoE Key Lab of Artificial Intelligence, AI Institute, Shanghai Jiao Tong University, China
} 

\begin{document}
\maketitle
\begin{abstract}
Crowd segmentation is a fundamental task serving as the basis of crowded scene analysis, and it is highly desirable to obtain refined pixel-level segmentation maps.
However, it remains a challenging problem, as existing approaches either require dense pixel-level annotations to train deep learning models or merely produce rough segmentation maps from optical or particle flows with physical models.
In this paper, we propose the Motion Prior-Aware Siamese Network (MPASNET) for unsupervised crowd semantic segmentation. 
This model not only eliminates the need for annotation but also yields high-quality segmentation maps.
Specially, we first analyze the coherent motion patterns across the frames and then apply a circular region merging strategy on the collective particles to generate pseudo-labels.
Moreover, we equip MPASNET with siamese branches for augmentation-invariant regularization and siamese feature aggregation.
Experiments over benchmark datasets indicate that our model outperforms the state-of-the-arts by more than 12\% in terms of mIoU.
\end{abstract}
\section{Introduction}
\begin{figure}
    \centering
    \includegraphics[width=\linewidth]{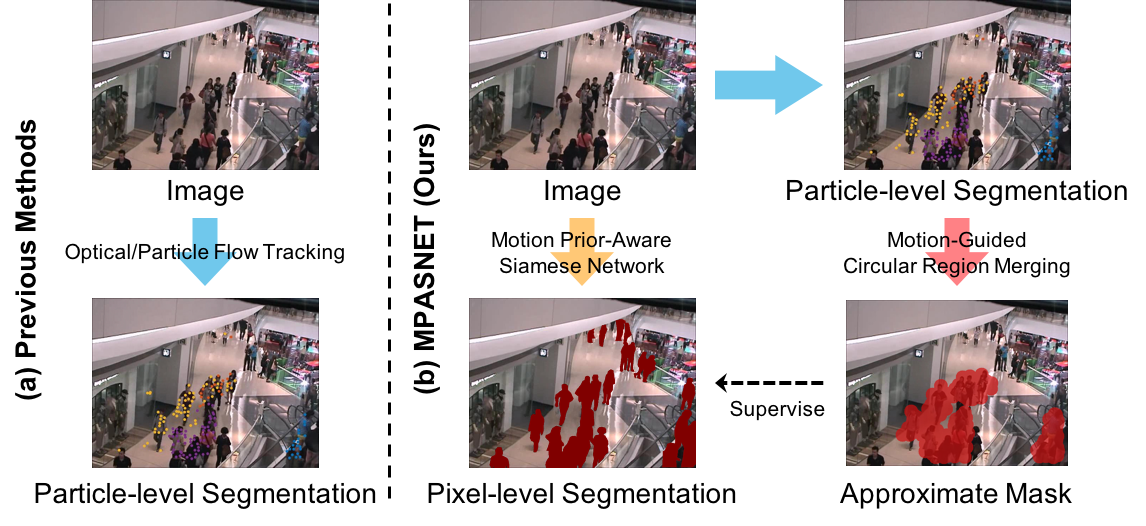}
    \caption{MPASNET in comparison with previous methods.}
    \label{fig:motivation}
\end{figure}

With the rapid growth of the urban population, crowded scene understanding~\cite{li2014crowded} from surveillance videos has attracted considerable attention in the computer vision community.
In particular, accurate pixel-level segmentation of the crowd is highly desirable to facilitate further analysis and forecasting of crowd behaviors.
To analyze the crowd pattern segmentation, a number of remarkable works have emerged from physical or statistical perspectives.
These works can be roughly categorized into three common types: (1) the flow field-based models~\cite{ali2007lagrangian,mehran2010streakline}, (2) the motion similarity-based models~\cite{zhou2013measuring,shao2014scene}, and (3) the deep learning-based models~\cite{kang2014fully}.
The flow field-based models treat the crowd pattern as time-varying flow fields, which are especially suitable for high-density complex scenes but usually fail in low-density or unstructured scenes.
Differently, the motion similarity-based models trace the long-term trajectories (tracklets) of keypoints (particles) and then group the motion particles into clusters with similarity metrics.
This kind of method generalizes better across different scenes compared to the flow field-based models, but they can only produce rough segmentation of particles as depicted in Fig.~\ref{fig:motivation}(a).
To conclude, neither of the above physical models can consistently obtain pixel-level refined segmentation results in various scenarios.
In contrast, recent years have witnessed the success of deep learning on pixel-level vision problems~\cite{long2015fully,zhao2017pyramid,wang2020deep}, yet few works~\cite{kang2014fully} have been done on deep crowd segmentation, mainly attributed to the prohibitive cost of exhaustive pixel-level annotation.

For the sake of a more accurate and generic model, we propose the Motion Prior-Aware Siamese Network (MPASNET) as an unsupervised deep learning model for crowd segmentation. 
The goal of MPASNET is to acquire transferable knowledge from a variety of unlabeled crowded scenes, thereby yielding high-quality pixel-level segmentation maps and meanwhile enhancing the cross-scene generalization.
To this end, we incorporate the motion similarity-based methods into the training paradigm of deep semantic segmentation models (as illustrated in Fig.~\ref{fig:motivation}(b)).
\cite{cheng2019online} has proposed a reinforcement learning framework that utilizes spatial-temporal Hessian matrix~\cite{yang2014large} and morphological operations to derive simple semantics as the environment information for better motion similarity measurement.
Differently, we take the spatial information of motion particles as the priors to construct approximate masks to train the deep segmentation models.
We will elaborate on the details in Sec.~\ref{motion}.
\begin{figure*}[t]
    \centering
    \includegraphics[width=\linewidth]{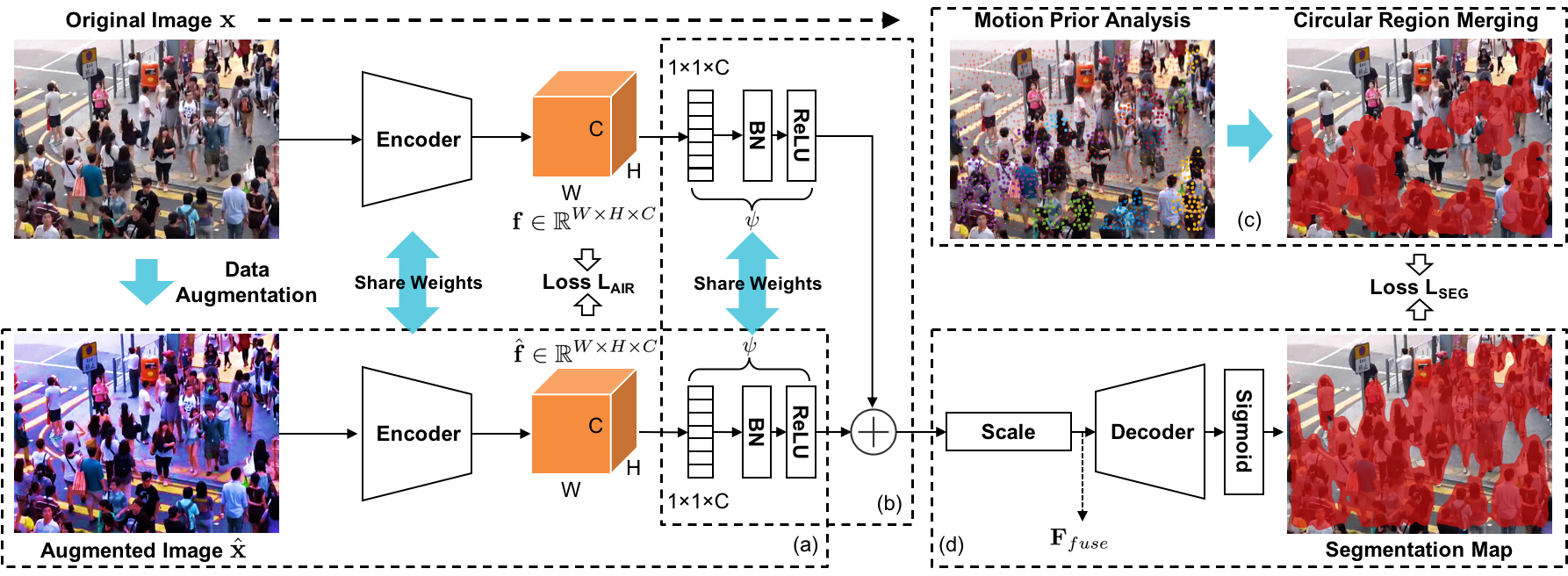}
    \caption{Illustrating the proposed MPASNET.
    (a) The siamese branch which takes the augmented images as the input.
    (b) Siamese feature aggregation.
    (c) Pseudo-labeling with motion-guided circular region merging.
    (d) The segmentation head.
    }
    \label{fig:framework}
\end{figure*}

It is also not trivial to devise the network structure and the loss function of MPASNET, due to the potential noise and error in the approximate segmentation labels.
As shown in Fig.~\ref{fig:framework}, we establish a siamese network~\cite{bertinetto2016fully,ye2019unsupervised} with bilateral branches, which respectively takes the original/augmented image as the input.
The proposed structure not only enables the feature aggregation between the original image the augmented image in the latent space for robust representation, but also serves as an augmentation-invariant regularization ($L_{AIR}$) on the encoder.
The foreground (crowds) and the background (others) tend to be highly-imbalanced in most crowded scenes, especially in low-density cases.
Besides, it is overconfident and irrational to supervise the model strictly using pixel-wise loss function with inaccurate ground truths.
Intuitively, we introduce a region-based criterion widely used in medical image segmentation, the Dice loss~\cite{milletari2016v,drozdzal2016importance}, which achieves substantial advantages in our experiments.

To our knowledge, this work is the first unsupervised deep learning-based model that focuses on crowd segmentation.
\textbf{Our contributions} are summarized as follows:
\textbf{(1)} We revisit the motion similarity-based methods and propose to produce approximate annotations for deep crowd segmentation with Circular Region Merging of masks centered at collective motion particles.
\textbf{(2)} We devise an end-to-end siamese network and the associated loss functions to learn from the self-produced pseudo-labels.
\textbf{(3)} We evaluate our unsupervised method on two representative datasets and significantly outperforms the state-of-the-art methods.
\section{Methods}
\subsection{Overview}
The overview of MPASNET is illustrated in Fig.~\ref{fig:framework}.
In brief, we generate pseudo-labels by leveraging the spatial information of collective particles as motion priors and then train a siamese network with the aid of pseudo labels.
During inference, the prediction head discards the siamese branch for efficiency and outputs segmentation maps without siamese feature aggregation.
Detailed descriptions of the pseudo-labeling algorithm (Fig.~\ref{fig:framework} (c)) and the siamese network (Fig.~\ref{fig:framework} (a)(b)) will be covered in Sec.~\ref{motion} and Sec.~\ref{siamese} respectively.

\subsection{Motion-Guided Circular Region Merging}\label{motion}
Surveillance videos are typically captured by fixed cameras; hence the background pixels are generally stationary.
On the other hand, the crowds are inclined to merge as self-organized dynamic systems with constituent spatially coherent motion patterns~\cite{moussaid2010walking}.
Therefore, we try to seek spatial cues from the motion of particles to outline a sketch of the crowd. 
A straightforward alternative is using the KLT tracker~\cite{shi1994good} to capture and track the keypoints as motion particles.
Whereas, many of the particles are irrelevant to the crowd semantics.
To remedy this deficiency, we adopt the collectiveness~\cite{zhou2013measuring} as the motion-similarity metric to remove outliers and noise.

At time $t$, for each particle $j$ in the $K$-nearest-neighbors of $i$, let $C_{t}(i, j)$ denote the velocity correlation between the $i$ and $j$, the weighted adjacency matrix of the crowd graph $\mathcal{C}$ is
\begin{equation}
\mathbf{W}_{t}[i, j]=\max \left(C_{t}(i, j), 0\right)~.
\end{equation}
Let $p^{(L)}_{i,j}$ denote a path of length $L$ that transitions for $L$ nodes from $i$ to $j$ with the trace of $\left\{n^{p}_{0} \rightarrow n^{p}_{1} \rightarrow \ldots \rightarrow p_{l}^{p}\right\}$, and let $\mathcal{P}^{(L)}$ denote all the possible paths of length $L$, the $L$-length path similarity is defined by
\begin{equation}
\label{l-path}
\nu_{L}[i, j]=\sum_{p^{(L)}_{i,j} \in \mathcal{P}^{(L)}}\prod_{k=0}^{L-1} \mathbf{W}_{t}\left[n^{(p)}_{k}, n^{(p)}_{k+1}\right]~.
\end{equation}
According to the algebraic graph theory~\cite{godsil2001algebraic}, $\nu_{L}= \mathbf{W}_t^{L}$, the $L$-th power of $\mathbf{W}_t$.
Weighted by an exponential decay factor $z$, the resulting path similarity of the graph can be written as
\begin{equation}
\mathbf{Z}_t[i, j]=\sum_{L=1}^{\infty} z^{L} \nu_{L}[i, j]~.
\end{equation}
Previous work~\cite{zhou2013measuring} has proven that
\begin{equation}\label{z}
\mathbf{Z}_t=(\mathbf{I}-z \mathbf{W}_t)^{-1}-\mathbf{I}~,
\end{equation}
and Eq.~\ref{z} converges when $0<z<\frac{1}{K}$.

The collectiveness of the $i$-th particle is defined by
\begin{equation}
    \phi_i=\sum_{j \in \mathcal{C}}\mathbf{Z}_t[i,j]~.
\end{equation}

\begin{figure}[t]
    \centering
    \includegraphics[width=\linewidth]{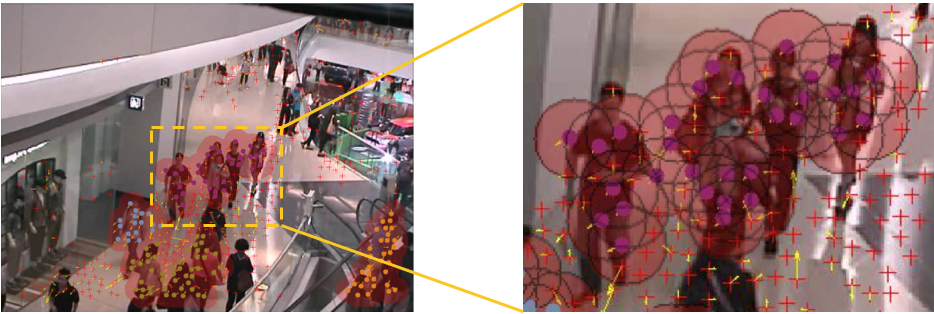}
    \caption{Motion-guided circular region merging. 
    The resulting mask is the merge of a collection of circular regions centered at the motion particles.
    Red crosses indicate the outliers detected by collectiveness thresholding, which are ignored.}
    \label{fig:circular}
\end{figure}

We then conduct collectiveness thresholding to eliminate outlier or noisy particles.
The upper bound of entries of $\mathbf{Z}$ is $\kappa=\frac{z}{1- zK}$~\cite{zhou2013measuring}.
We set the threshold to $0.6\kappa$.
The left of Fig.~\ref{fig:circular} shows that the outliers are detected and filtered effectively. 
Another notable observation is that the collective motion particles are scattered over most regions of the crowds.
In light of this, we utilize a simple yet effective approach to obtain pseudo segmentation maps in a bottom-up manner, which we call Circular Region Merging (CRM).
As shown in the right of Fig.~\ref{fig:circular}, we take each particle as the center of a circular region, and the final pseudo-mask is the merge of all the produced regions.
Since the movement of the crowd in consecutive frames is inconspicuous, we preserve only one of every 20 frames in practice.
\subsection{Motion Prior-Aware Siamese Network}\label{siamese}
The model structure is illustrated in Fig.~\ref{fig:framework}.
Since the pseudo-labels generated from the continuous frames merely provide a rough estimation of the segmentation maps, we suggest employing region-based rather than pixel-based loss functions.
To loosen the pixel-level supervision, here we introduce the Dice loss that has been widely used in medical image segmentation~\cite{milletari2016v,drozdzal2016importance} with Laplace smoothing.
Suppose the image size is $W^\prime\times H^\prime$, let $o_i \in (0,1)$ and $y_i \in \{0,1\}$ denote the $i$-th output and the associated pseudo-label, the segmentation loss is
\begin{equation}
L_{SEG}=\frac{1}{W^\prime\times H^\prime}\left[1-\frac{1+2 \sum_{i} o_{i} y_{i}}{1+\sum_{i} o_{i}+\sum_{i} y_{i}}\right]~.
\end{equation}
\noindent \textbf{Augmentation-invariant regularization.}
Data augmentation is a standard policy in deep learning for computer vision.
The latent feature of an augmented image is expected to be consistent with the counterpart of the original one~\cite{ye2019unsupervised,yang2020coarse}.
Let $\mathbf{F}\in \mathbb{R}^{W\times H\times C}$ and $\hat{\mathbf{F}}\in \mathbb{R}^{ W\times H\times C}$ be the intermediate feature of the original and augmented images (see Fig.~\ref{fig:framework}), we ensure this property by a mean squared error loss:
\begin{equation}
\small
    L_{AIR} = \frac{1}{W\times H \times C}\sum_{i=1}^{W}\sum_{j=1}^{H}\sum_{k=1}^{C} (\mathbf{F}[i,j,k]-\hat{\mathbf{F}}[i,j,k])^2~.
\end{equation}
\noindent \textbf{Siamese feature aggregation.}
As shown in Fig.~\ref{fig:framework}(b), the siamese features are aggregated in an additive attention manner to increase the input diversity of the decoder.
A weight-shared convolutional module $\psi$ (a $1\times1\times C$ convolutional layer followed by batch normalization and ReLU activation) is inserted between the encoder and the decoder.
The fused feature is calculated by $\mathbf{F}_{fuse}=(\psi(\mathbf{F})+\psi(\hat{\mathbf{F}}))/{2}$.
During inference, we simply use $\mathbf{F}_{fuse}=\psi(\mathbf{F})$. 
The segmentation head then processes the aggregated features to output segmentation maps.
The overall loss is defined as $L=L_{AIR}+ L_{SEG}$.

\section{Experiments}

\subsection{Experiment Settings}
\textbf{Datasets.}
The Collective Motion Database~\cite{zhou2013measuring} contains 413 video clips from 62 crowded scenes. 
The CUHK~\cite{shao2014scene} dataset includes 474 video clips from 215 scenes.
Following~\cite{cheng2019online}, we evaluate our method on 6 representative scenarios with a total of 20 clips.
They are (1) \texttt{2\_36}, (2) \texttt{1\_8groupSplit}, (3) \texttt{2\_000455712}, (4) \texttt{marathonround}, (5) \texttt{2\_35} and (6) \texttt{3\_1\_12}.
Since the segmentation ground truths are not publicly available, we annotate them with LabelMe~\cite{russell2008labelme}.

\noindent \textbf{Implementation details.}
We use $K=20$ and $z=\frac{1}{2K}$.
In CRM, we set the radius of circular regions to 20 pixels.
We utilize HRNetV2-W48~\cite{wang2020deep} pretrained on ImageNet~\cite{deng2009imagenet} as the encoder and a single convolutional module as the decoder.
The optimizer is SGD with the learning rate of 0.02, the momentum of 0.9, and the weight decay of $1\times 10^{-4}$.
We use \texttt{ColorJitter} and \texttt{RandomGrayScale} with default parameters in PyTorch~\cite{paszke2019pytorch} to get augmented images.

\subsection{Comparison with the State-of-the-Arts}
\begin{figure*}[ht]
    \centering
    \includegraphics[width=\linewidth]{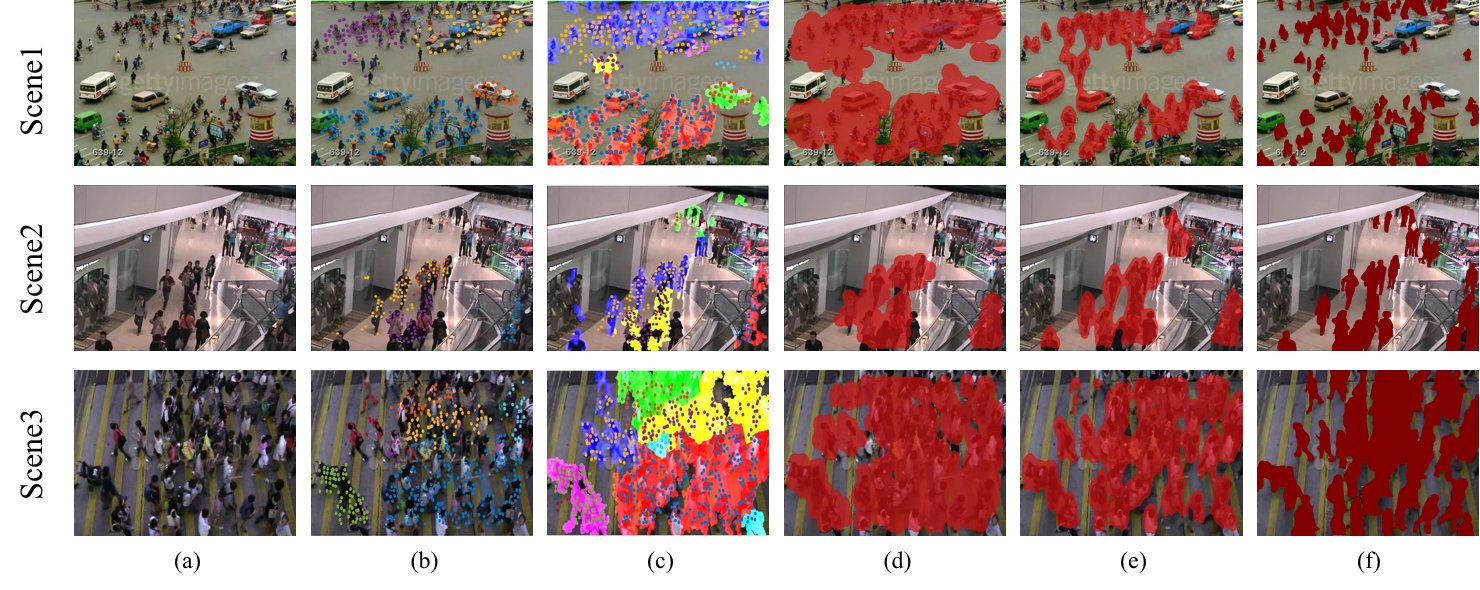}
    \caption{Qualitative comparison with state-of-the-art methods.
    (a) Original frames.
    (b) Motion particles extracted by CM~\cite{zhou2013measuring}.
    (c) Segmentation results of CrowdRL~\cite{cheng2019online}.
    (d) Pseudo-labels synthesized by CM~\cite{zhou2013measuring} with Circular Region Merging.
    (e) Segmentation results of MPASNET.
    (f) Ground truths.
    Our method achieves a clear improvement over others.} 
    \label{fig:visualization}
\end{figure*}

We adopt Intersection over Union (IoU) for evaluation and refer the mean IoU across all the evaluated scenes to as \texttt{mIoU}.
Tab.~\ref{tab:results} lists the IoU results on the 6 scenes and the overall mIoU, for each compared methods.
Since CM~\cite{zhou2013measuring} only conduct particle-level segmentation (see Fig.~\ref{fig:visualization}(b)), we enhance it with the proposed CRM for comparison.
As indicated in Tab.~\ref{tab:results}, MPASNET surpasses the state-of-the-arts by a large margin of more than 12\% in terms of cross-scene mIoU, proving both the effectiveness and the generalization ability of our method.
As shown in Fig.~\ref{fig:visualization}, our approach produces more refined segmentation maps compared with existing methods.
In aerial views, e.g. Scene1, our model is not the best.
The reason may be that the crowd is quite small while the receptive field of Convolutional Neural Networks is limited.
\begin{table}[t]
    \centering
    \footnotesize
    \caption{Quantitative comparison with state-of-the-art methods in terms of IoU. MPASNET* means training MPASNET without the siamese branch (i.e., without augmentation-invariant regularization and siamese feature aggregation).}
    \setlength{\tabcolsep}{1pt}
    \begin{tabular}{lccccccccccccccccc}\toprule
    \textbf{Method} &\textbf{Scene1}&\textbf{Scene2}&\textbf{Scene3}&\textbf{Scene4}&\textbf{Scene5}&\textbf{Scene6}&\textbf{mIoU}\\\midrule
    CM~\cite{zhou2013measuring} + CRM&0.2561&0.6410&0.5154&\textbf{0.6921}&0.5251&0.3608&0.4949\\
         CrowdRL~\cite{cheng2019online}&\textbf{0.6279}&0.4832&0.5746&0.4673&0.3298&0.6614&0.5213  \\
         MPASNET*& 0.4707&0.6547&0.6151&0.5749&0.6614&0.6685&0.6009\\
         MPASNET (Ours)& 0.4887&\textbf{0.7078}&\textbf{0.6617}&0.6553&\textbf{0.6655}&\textbf{0.6892}&\textbf{0.6447}\\\bottomrule
    \end{tabular}
    \label{tab:results}
\end{table}
\subsection{Ablation Study}
In this section, we investigate the effect of each components of our model on the final results.
Firstly, we start with the full model and then remove the Siamese Feature Aggregation (SFA) and Augmentation-Invariant Regularization (AIR) respectively.
As indicated in Tab.~\ref{tab:ablation}, both components facilitate the final performance.
Secondly, we replace Dice loss with the Cross Entropy loss (which was adopted in previous crowd segmentation method~\cite{kang2014fully}) for $L_{SEG}$ to justify our discussions in Sec.~\ref{siamese}.
As can be observed in Tab.~\ref{tab:ablation}, Dice loss achieves across-board improvements over the regular Cross Entropy loss.
In summary, all the investigated components result in a performance gain.
Particularly, AIR contributes the most.

\begin{table}[t]
    \centering
    \caption{Effect of each component. ``AIR'': Augmentation-Invariant Regularization. ``SFA'': Siamese Feature Aggregation. ``Dice'': Dice loss~\cite{drozdzal2016importance}. ``CE'': Cross Entropy loss~\cite{kang2014fully}.}
    \vspace{-17pt}
    \setlength{\tabcolsep}{4.5pt}
    \begin{tabular}{cccccccc}\\\toprule
         \textbf{AIR}&\textbf{SFA}&\textbf{Loss}&\textbf{Scene1}&\textbf{Scene2}&\textbf{Scene3}&\textbf{mIoU}\\\midrule
         \checkmark&\checkmark&Dice& \textbf{0.4887}&\textbf{0.7078}&\textbf{0.6617}&\textbf{0.6194}\\
         \checkmark&&Dice&0.4709&0.6856&0.6462&0.6009\\
         &\checkmark&Dice&0.4638&0.6788&0.6248&0.5891\\
         \checkmark&\checkmark&CE&0.4735&0.6750&0.6489&0.5991\\\bottomrule
    \end{tabular}
    \label{tab:ablation}
\end{table}


\section{Conclusion}
In this paper, we propose the Motion Prior-Aware Siamese Network for unsupervised deep crowd semantic segmentation.
Different from existing methods, our model takes advantage of the motion priors to produce pseudo-labels without human effort, thus learning to generate high-quality segmentation maps in the absence of ground truth.
Experiments and ablation studies show that our framework achieves substantial improvements over the state-of-the-art unsupervised methods. 
\section{REFERENCES}
\bibliographystyle{IEEEbib}

\end{document}